\documentclass[letterpaper, 10 pt, conference]{ieeeconf}  

\IEEEoverridecommandlockouts                              

\usepackage{soul}
\usepackage{multirow}
\usepackage{graphicx}
\usepackage{amsmath,amssymb,amsfonts}
\usepackage{svg}
\usepackage{multirow}
\usepackage{hyperref}
\usepackage{pgfplots}
\usepackage[skip=1ex, font=small]{caption}

\usepackage{booktabs}  
\usepackage[skip=1ex, font=small]{caption}

\usepackage{color}
\usepackage[normalem]{ulem} 
\newcommand{\red}[1]{\textcolor{red}{#1}}

\newcommand{\green}[1]{\textcolor{green}{#1}}

\title{\LARGE \bf Semantics-aware LiDAR-Only Pseudo Point Cloud Generation \\for 3D Object Detection}

\author{Tiago Cortinhal$^{1,*}$, Idriss Gouigah$^{1,*}$, and Eren Erdal Aksoy$^{1}$ 
\thanks{*These authors contributed equally to this work.}
\thanks{$^{1}$Halmstad University, School of Information Technology,
        Center for Applied Intelligent Systems Research, Halmstad, Sweden
      {\tt\small {\{tiago.cortinhal, idriss.gouigah\}@hh.se}}%
    }
\thanks{This work was funded by the European Union (grant no. 101069576).
Views and opinions expressed are however those of the author(s) only and do not necessarily reflect those of the European Union or the European Climate, Infrastructure and Environment Executive Agency (CINEA). Neither the European Union nor the granting authority can be held responsible for them.}
\thanks{Special thanks to Abu Mohammed Raisuddin for his time helping with this paper.}
}

\makeatletter
\let\@oldmaketitle\@maketitle
\renewcommand{\@maketitle}{\@oldmaketitle
\centering 
    \includegraphics[width=\textwidth]{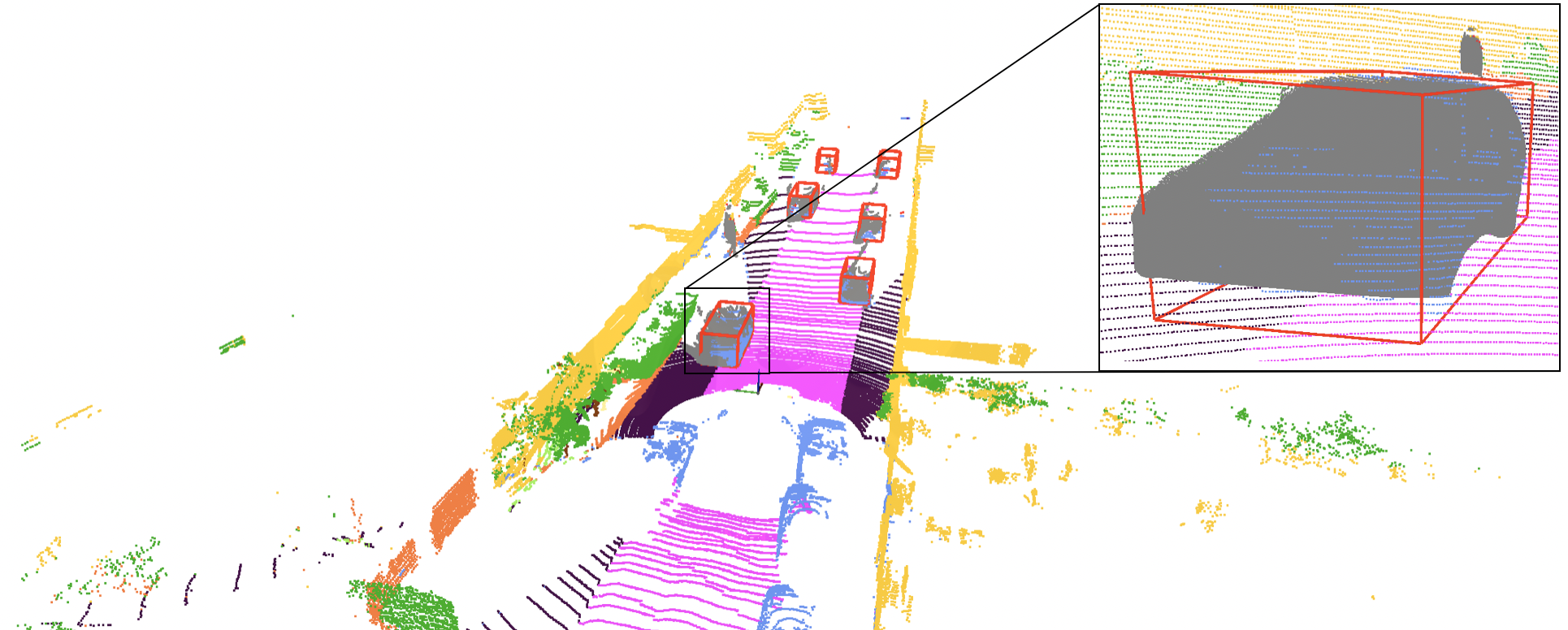}
    \captionof{figure}{ 
    We propose a novel modular framework that augments raw and sparse LiDAR scans with denser pseudo point cloud data by solely relying on the LiDAR sensor and the scene semantics without requiring any additional modality, such as cameras.
    The final output is semantically segmented dense full-scan point clouds with 3D bounding boxes for each detected object (e.g., cars) in the scene. 
    In the zoomed image, the generated dense pseudo point cloud for the car is shown in gray, whereas the original sparse LiDAR data for the very same car object is indicated in blue. Note that objects are detected only for the front view. Best seen in color.
    }
    \label{fig:scene}
   }

\begin{document}

\maketitle
\thispagestyle{empty}
\pagestyle{empty}

\begin{abstract}
Although LiDAR sensors are crucial for autonomous systems due to providing precise depth information, they struggle with capturing fine object details, especially at a distance, due to sparse and non-uniform data. 
Recent advances introduced \textit{pseudo-LiDAR}, i.e., synthetic dense point clouds, using additional modalities such as cameras to enhance 3D object detection.
We present a novel LiDAR-only framework that augments raw scans with denser pseudo point clouds by solely relying on LiDAR sensors and scene semantics, omitting the need for cameras. 
Our framework first utilizes a segmentation model to extract scene semantics from raw point clouds, and then employs a multi-modal domain translator to generate synthetic image segments and depth cues without real cameras. 
This yields a dense pseudo point cloud enriched with semantic information.
We also introduce a new semantically guided projection method, which enhances detection performance by retaining only relevant pseudo points.
We applied our framework to different advanced 3D object detection methods and reported up to $2.9\%$ performance upgrade. 
We also obtained comparable results on the KITTI 3D object detection test set, in contrast to other state-of-the-art LiDAR-only detectors.

\end{abstract}

\section{INTRODUCTION}

Recent works on LiDAR-based perception show that the LiDAR modality plays a pivotal role in enabling autonomous systems to understand complex environments. 
While LiDARs excel in capturing precise depth information and generating accurate point cloud data in a wide field of view, these active sensors may encounter limitations in capturing fine-grained object details, particularly for distant or unreflective surfaces. 
This is mainly because the rendered LiDAR data is sparse, unstructured, and follows a non-uniform sampling. 

LiDAR-only 3D object detection techniques \cite{Lang_2019_CVPR,mao2021pyramid,Sheng_2021_ICCV,Shi_2019_CVPR,9018080,Yang_2020_CVPR,Yang_2019_ICCV,8886046} conventionally rely on such raw sparse LiDAR data to perceive the surroundings. 
Recent scientific investigations~\cite{wang2019pseudo,you2020pseudo,10.1007/978-3-030-58601-0_19,VirConv}  reveal the potential of \textit{pseudo-LiDAR}, a novel technique that leverages mono and stereo imaging data to generate \textit{synthetic point cloud representations}.
These studies heavily rely on the fusion of pseudo-LiDAR data with the original LiDAR scans. 
Such a fusion of both data sources indeed augments the information available to the 3D object detection system. 
This is mainly because, unlike the original sparse LiDAR point clouds, the pseudo point clouds are relatively denser, enhancing the overall richness of the data. 
Notably, the synthetic data from pseudo-LiDAR fills the gaps left by traditional LiDAR scans. 
By capitalizing on the complementary strengths of both data sources, the integrated approaches demonstrate substantial potential in enhancing the object detection. 

With this motivation, we introduce a novel modular framework that augments raw LiDAR scans with denser pseudo point cloud data. 
Our proposed framework differs from all other relevant works~\cite{wang2019pseudo,you2020pseudo,10.1007/978-3-030-58601-0_19,VirConv} in that it solely relies on the LiDAR sensor and the scene semantics without incorporating any additional modality such as mono or stereo cameras. Furthermore, our framework is unique since the final output is semantically segmented dense point clouds with 3D bounding boxes for each detected object in the scene as shown in Fig.~\ref{fig:scene}.

As the first step, our framework employs an off-the-shelf segmentation model (e.g., SalsaNext~\cite{10.1007/978-3-030-64559-5_16}) to extract the scene semantics of the raw and sparse \textit{full-scan} LiDAR point clouds. 
Next, a state-of-the-art multi-modal domain translator, named TITAN-Next~\cite{2023arXiv230207661C}, is triggered to translate LiDAR segments into \textit{expected synthetic image segments and depth cues}, without requiring any real camera data. 
A dense pseudo point cloud enriched with semantic information is then directly rendered from these estimated synthetic camera depth cues. 

The higher the density of pseudo points, the greater the level of object details in the scene, but also the larger the noise in the point cloud and the longer the computation time.
The inclusion of all new pseudo points may lead to an excessive number of irrelevant and redundant points, potentially overwhelming, for instance, the object detection system and hindering its efficiency~\cite{VirConv}. 
Therefore, in the context of point cloud augmentation, analysis of this factor is of utmost importance as the amount of density introduces challenges to the downstream tasks.  
To address this issue, we introduce a novel Semantically Guided Projection (SGP) technique to select the most important foreground points in the pseudo point cloud.
This approach selectively retains points only from these classes that are highly relevant to the object detection task (such as pedestrians, vehicles, or cyclists), thus, significantly reducing the computational burden and improving the overall detection performance.
This proposed targeted projection not only reduces the data volume but also enhances the discriminative power of the pseudo point cloud for object detection.

We applied our framework to different advanced 3D object detection methods (e.g., PointPillars~\cite{lang2019pointpillars}, SECOND~\cite{yan2018second}, and Voxel-RCNN~\cite{deng2021voxel}) and obtained a certain performance upgrade with little to no modification to the detection model. These experiments convey that our framework is agnostic to object detector architectures, however, works with higher performance in the case of having detectors specifically designed for pseudo point cloud data, such as VirConv~\cite{VirConv}.
We also obtained comparable results on the KITTI test set in contrast to other state-of-the-art LiDAR-only detectors.

In a nutshell, our contributions are manifold:
\begin{itemize}
\item We propose a modular LiDAR-only pseudo point cloud generation framework without requiring any additional modality, such as cameras.
\item We introduce a new projection scheme, SGD, to reduce the density of the pseudo point cloud by selecting the points with the most relevant semantic information.
\item Our framework returns not only dense point clouds but also semantic segments and 3D object bounding boxes. 
\item We conduct extensive experiments on different advanced 3D object detection models and show the impact of our synthesized pseudo point cloud data.
\end{itemize}

\setcounter{figure}{1}
 \begin{figure*}[ht]
    \centering
    \resizebox{\textwidth}{!}{%
    \includegraphics{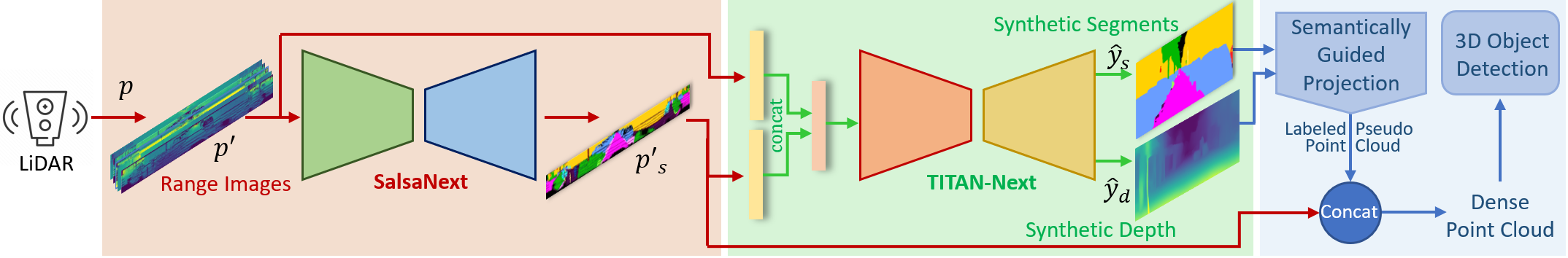}
    }
    \caption{
    Our proposed modular framework has three blocks, each of which is depicted by a unique background color. 
    In the red box, a raw 3D LiDAR point cloud $p$ is first projected onto the 2D range image plane $p^{\prime}$ to be segmented by \textit{SalsaNext}~\cite{10.1007/978-3-030-64559-5_16}, predicting $p^{\prime}_{s}$.
    The green box highlights the  TITAN-Next~\cite{2023arXiv230207661C} generative model, conditioned on the concatenated $p^{\prime}$ and $p^{\prime}_{s}$ to generate a synthetic segment map $\widehat{y}_{s}$ and a synthetic depth map $\widehat{y}_{d}$ in the expected front-view camera space, without requiring any camera information.   
    Finally, as shown in the blue box, these generated segment and depth maps ($\widehat{y}_{s}$ and $\widehat{y}_{d}$)  are synthesized to render a labeled pseudo point cloud by our new Semantically Guided Projection method.
    The segmented original LiDAR point cloud ($p^{\prime}_{s}$) is then concatenated with the pseudo counterpart to obtain the final semantically segmented dense point cloud, which is further fed to the subsequent 3D object detector.
    }
    \label{fig:pipeline}
\end{figure*}

\section{Related Work}
Several studies dedicated to 3D object detection employ point cloud augmentations, however, they are mostly multimodal.
In such methods, information coming from other sensors (e.g., RGB cameras) is fused at the early stage to enhance the raw LiDAR point clouds.  
These early fusion methods can be divided into two categories: Point painting~\cite{vora2020pointpainting, xu2021fusionpainting, wang2021pointaugmenting, huang2020epnet} and pseudo point cloud generation~\cite{wang2019pseudo, you2020pseudo, VirConv, yin2021multimodal,wu2022sparse}.

\paragraph{Point Painting}
PointPainting \cite{vora2020pointpainting} is a pioneering method that uses camera images to \emph{paint} 3D LiDAR point clouds. 
PointPainting effectively leverages the high-resolution and semantic details from cameras to improve object localization and classification in dense and sparse LiDAR data. 
This way, the model achieves superior performance in detecting objects with fine-grained details and can handle occlusions, making it well-suited for complex urban driving scenarios. 
Following methods~\cite{xu2021fusionpainting, wang2021pointaugmenting} extend the idea by painting the point cloud using more relevant features. 

\paragraph{Pseudo point cloud generation}
These methods such as MVP~\cite{yin2021multimodal} leverage the camera information by generating a depth map from an image (either by disparity map or using a neural network), which is then projected into the LiDAR space. The goal is to reduce the sparsity of the LiDAR data by increasing the overall number of points. 
Sparse Fuse Dense~\cite{wu2022sparse} decorates the generated pseudo point cloud with RGB values from the camera and uses grid-wise attentive fusion to merge features coming from the pseudo point cloud with that of the raw point cloud data. 
More recently, VirConv~\cite{VirConv} introduced a model specifically designed to tackle the drawbacks of pseudo point clouds, such as the longtail and misalignment issues. 
With noise-resistant convolution and stochastic voxel discard, VirConv~\cite{VirConv} manages to reach a high level of performance in the car detection challenge of the KITTI dataset. 

Although these methods in both mainstream categories help improve the performance of 3D detection, they require an additional modality such as RGB cameras, and come with the cost of high computation time. 
Moreover, to the best of our knowledge, these two categories have so far been studied separately. 
Our approach differs from all these methods as it couples both point painting and pseudo point cloud generation approaches, while we solely rely on the LiDAR sensor and employ the semantic information to select the most valuable pseudo points, reducing the processing cost.
 
\section{Method}

As illustrated in Fig.~\ref{fig:pipeline}, we proposed a framework involving three individual modules. The blue box represents the main technical contribution of this work, whereas the red and green boxes involve other trained state-of-the-art networks.

\subsection{Pipeline}
As shown in the red box in Fig.~\ref{fig:pipeline}, the proposed pipeline starts with the semantic segmentation of the raw and sparse \textit{full-scan} LiDAR point cloud. 
Given a point cloud $p\in \mathcal{R}^{N \times 4}$, where $N$ is the number of points with $x$, $y$, $z$ point coordinates and $i$ intensity values, we first employ an off-the-shelf semantic segmentation model (e.g., SalsaNext~\cite{10.1007/978-3-030-64559-5_16}) to extract the scene semantics. In this case, the point cloud $p$ is first projected onto the 2D range image plane $p^{\prime}$ to be segmented by \textit{SalsaNext}~\cite{10.1007/978-3-030-64559-5_16}, predicting a semantic segment map $p'_s \in \mathcal{R}^N$.
Please note that any other segmentation model can be employed here instead of SalsaNext~\cite{10.1007/978-3-030-64559-5_16}.

At the next step (highlighted in the green box), the concatenated $p^{\prime}$ and $p^{\prime}_{s}$ are fed to a generative domain translation module, named TITAN-Next~\cite{2023arXiv230207661C},  to generate a synthetic segmentation map $\widehat{y}_{s}$ and a synthetic depth map $\widehat{y}_{d}$ in the expected front-view camera space, without requiring a real camera. TITAN-Next is a multi-modal domain translation network~\cite{Cortinhal_2021_ICCV} and maps the segmented LiDAR range view projection onto the expected camera image space. Thus, TITAN-Next synthesizes semantic segments and the corresponding depth information in this estimated camera space even though no real camera is available. 
We here again note that since TITAN-Next is not forming the main contribution of this framework, it can be replaced by any other cross-domain mapping method working in a multi-modal setup.

Finally, as depicted in the blue box in Fig.~\ref{fig:pipeline}, these generated segment and depth maps ($\widehat{y}_{s}$ and $\widehat{y}_{d}$)  are synthesized to render a labeled pseudo point cloud by our new Semantically Guided Projection method, which is detailed next.

\subsection{Semantically Guided Projection (SGP)}
Methods such as~\cite{wang2019pseudo} generate a pseudo point cloud from depth estimation and feed it as input to the detector entirely. 
However, the density of the generated point cloud makes the computation time prohibitive for real-time applications. 
Methods such as MVP~\cite{yin2021multimodal} reduce the size of the point cloud by leveraging 2D detection results, such that a point cloud is created only in the regions of interest yielded by a 2D detector.
However, the 2D detection results are redundant together with the 3D detection and cannot be leveraged in other downstream tasks. 
Instead, we propose to leverage the semantic segmentation information in the LiDAR space to reduce the size of the pseudo point cloud, while still being useful for other relevant subsequent downstream tasks such as free space detection, in addition to object detection.

\begin{figure}[!t]
    \centering
    \includegraphics[width=0.5\textwidth]{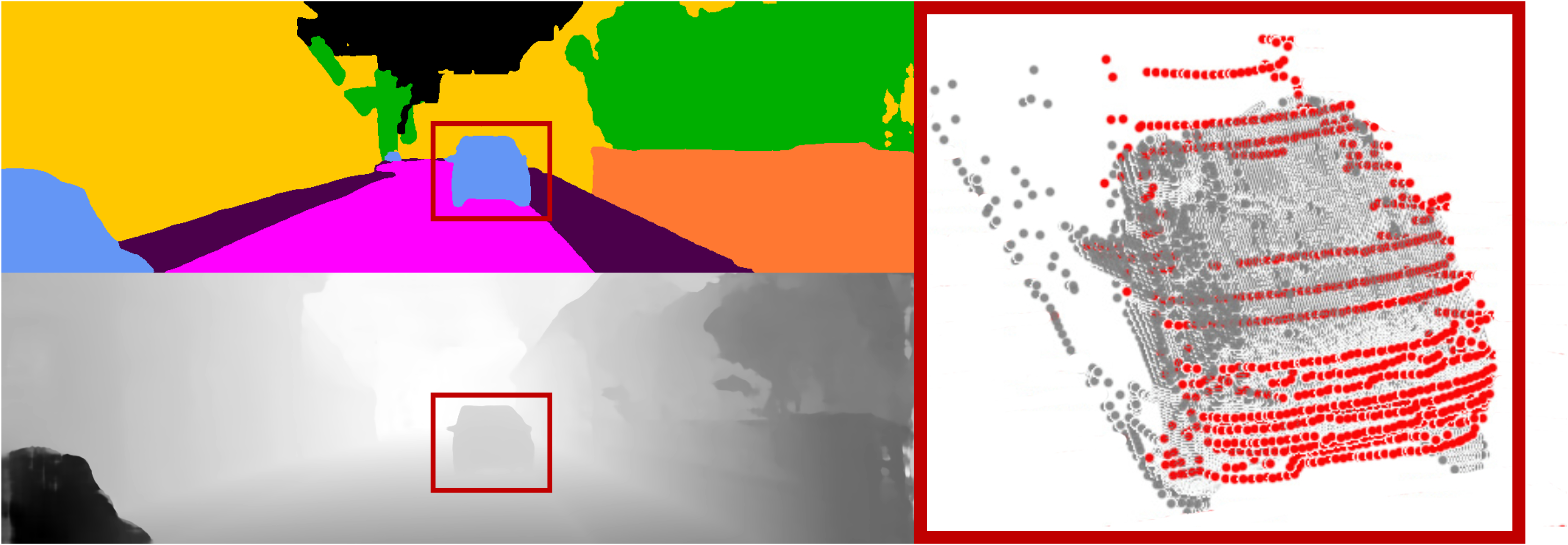}
    \caption{Semantically Guided Projection. The synthetic segment (top-left) and depth (bottom-left) maps are pixel-to-pixel associated only for the relevant semantic classes. In this example, the dense pseudo point cloud shown on the right is generated only for the segmented car object. Note that red points are from the original LiDAR point cloud, whereas gray points represent the generated pseudo points.}
    \label{fig:sgp}
\end{figure}

Semantic Guided Projection (SGP) is a very simple yet efficient method. 
SPG translates the rendered semantic segmentation maps ($\widehat{y}_{s}$)  and depth estimation ($\widehat{y}_{d}$)  into 3D dense pseudo point cloud data and filters out noisy points as illustrated in Fig.~\ref{fig:sgp}. 

First, SPG associates $\widehat{y}_{s}$ and $\widehat{y}_{d}$  pixel to pixel since both maps are aligned.  
Here, SPG only selects depth values associated with pertinent object classes for object detection tasks, such as cars, pedestrians, or cyclists.
This vastly reduces the density of the generated pseudo point cloud. 
Next, by utilizing the available calibration information, SGP conducts a transformation that projects the selected depth estimations from the camera's intrinsic frame of reference to the LiDAR's extrinsic frame of reference.
The correspondence between the depth points and the 3D spatial points can be mathematically described as follows:
\begin{equation}
    \Tilde{y} = (u \times z, v \times z, z, 1)~~,
\end{equation}
\begin{equation}
    T^{velo}_{cam} = (T^{cam}_{velo})^{-1}~~,
\end{equation}
\begin{equation}
    (T^{cam}_{velo}) \times \Tilde{y} = \Tilde{x}~~,
\end{equation}
where $T^{cam}_{velo}$ is the projection matrix from 3D LiDAR space to 2D camera $(u,v)$ coordinate space, and $\Tilde{y}$ is the homogenous coordinates with $z$ being the predicted depth.

As comprehensively described in~\cite{VirConv,yin2019fusionmapping}, pseudo point clouds introduce inherent challenges compared to real counterparts, primarily due to inaccurate depth estimation around object boundaries, leading to misalignment and long tails, particularly for near object edges. 
While specialized methods such as VirConv~\cite{VirConv} address these concerns, not all techniques (including PointPillars~\cite{Lang_2019_CVPR}) are optimized for pseudo point clouds, resulting in noise and hindering the performance of the off-the-shelf object detectors.

\begin{figure}[]
    \centering
    \includegraphics[width=0.4\textwidth]{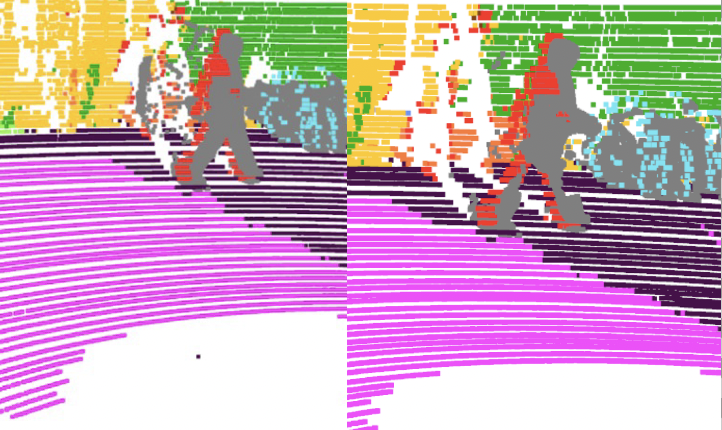}
    \caption{On the right, we show the effects of applying our cleaning to the pseudo point cloud. Notice how the pseudo points (in gray) appear behind the person on the left are properly removed, and we are left with pseudo points in proximity to real points corresponding to the segmented person.}
    \label{fig:clean_person}
\end{figure}

To tackle this issue, we enrich SGP with a cleaning strategy rooted in the analysis of original sparse LiDAR data. The core rationale here is that objects in the real world are expected to yield corresponding LiDAR data points in their vicinity.
Given that LiDAR scans provide a relatively coarse representation of the scene, we utilize this insight to identify outliers within the pseudo point cloud. 
Specifically, if a pseudo point lacks nearby real LiDAR points, it is indicative of an absence of a corresponding physical object in the real world. Consequently, we establish a volumetric region around each pseudo point and eliminate those lacking real points within this specific volume. This methodology ensures the selection of only those pseudo points that closely correspond to objects, effectively mitigating the challenges associated with long tail and ghost measurements caused by border effects in semantic segmentation and depth prediction. Fig.~\ref{fig:clean_person} shows a sample scene before and after applying our pseudo point cloud cleaning method. 


\subsection{3D Object Detection}

Finally, the segmented original LiDAR point cloud ($p^{\prime}_{s}$) is then concatenated with the cleaned pseudo counterpart to obtain the final semantically segmented dense point cloud, which is fed to the subsequent 3D object detector as shown in the blue box in Fig.~\ref{fig:pipeline}.

We here note that our method aims at augmenting the point cloud to improve the detection results while bringing little to no modification to the detection model.
However, after extensive investigation, it has become clear that to be processed correctly and yield better results, a pseudo point cloud needs to be processed by a dedicated model. 
For instance, VirConv~\cite{VirConv} constructed from Voxel-RCNN~\cite{deng2021voxel} introduces a specific virtual sparse convolution to deal with the challenges coming with the pseudo point clouds. 
VirConv~\cite{VirConv} makes use of a set of new modules called Noise Resistant Convolution (NRConv) and  Stochastic Voxel Discard (StVD) to improve the detection results by a large margin. 
Thus, we perform experiments with both generic object detectors (e.g., PointPillars~\cite{Lang_2019_CVPR})  and object detectors designed for pseudo point clouds (e.g., VirConv~\cite{VirConv}). 



\section{Datasets \& Results}
We here present our results on the KITTI datasets and compare with other state-of-the-art methods in the literature. 

\subsection{The KITTI dataset}
The KITTI 3D Object Detection dataset \cite{kittidataset} is a widely used benchmark for autonomous driving. It includes 7,481 annotated training frames and 7,518 test frames captured in real-world urban environments. The dataset comprises various sensor data modalities, including high-resolution RGB images, 3D point clouds from LiDAR sensors, and calibration data for sensor alignment. It offers precise 3D bounding box annotations for objects like cars, pedestrians, and cyclists in each frame.

\begin{table*}[!t]
    \centering
\resizebox{\textwidth}{!}{
  \begin{tabular}{|l|l|l|l|l|l|l|l|l|l|l|l|l|l|}
    \hline
    \multirow{2}{*}{} &
      \multicolumn{3}{c|}{Car} & 
      \multicolumn{3}{c|}{Pedestrian} & 
      \multicolumn{3}{c|}{Cyclists} & 
      \multicolumn{3}{c|}{Overall} \\
      \hline
    & Easy & Med & Hard & Easy & Med & Hard & Easy & Med & Hard & Easy & Med & Hard   \\
    \hline
   PointPillars~\cite{Lang_2019_CVPR} & 87.5 & 78.7 & 75.7 & 57.2 & 50.9 & 46.3 & 82.5 & 62.0 & 58.6 & 75.7 & 63.9 & 60.2 \\ 
   Augmented PointPillars  (Ours) & 88.3 & 79.5 & 76.5 & 58.9 & 52.6 & 47.9 & 82.7 & 63.1 & 59.7 & 76.6 & 65.1 & 61.4 \\ 
   Delta & \green{+0.8} & \green{+0.8} & \green{+0.8} & \green{+1.7} & \green{+1.7} & \green{+1.6} & \green{+0.2} & \green{+1.1} & \green{+1.1} & \green{+0.9} & \green{+1.2} & \green{+1.2} \\ \hline
SECOND~\cite{yan2018second} & 88.1 & 78.2 & 73.2 & 60.0 & 52.8 & 47.3 & 75.8 & 61.1 & 57.5 & 74.6 & 64.0 & 59.4 \\ 
Augmented SECOND  (Ours) & 88.7 & 78.7 & 75.1 & 63.1 & 54.7 & 48.4 & 79.8 & 62.5 & 59.3 & 77.2 & 65.3 & 60.9 \\ 
Delta & \green{+0.6} & \green{+0.5} & \green{+1.9} & \green{+3.1} & \green{+1.9} & \green{+1.1} & \green{+4} & \green{+1.4} & \green{+1.8} & \green{+2.6} & \green{+1.3} & \green{+1.5} \\
    \hline
Voxel-RCNN~\cite{deng2021voxel} & 92.0 & 84.9 & 82.6 & - & - & - & - & - & - & - & - & - \\
Augmented Voxel-RCNN (Ours)  & 92.8 & 85.7 & 83.3 & - & - & - & - & - & - & - & - & - \\
Delta & \green{+0.8} & \green{+0.8} & \green{+0.7} & - & - & - & - & - & - & - & - & - \\
Augmented Voxel-RCNN (Ours) + NRConv & 92.6 & 87.8 & 85.4 & - & - & - & - & - & - & - & - & - \\
Delta & \green{+0.6} & \green{+2.9} & \green{+2.8} & - & - & - & - & - & - & - & - & - \\ \hline
  \end{tabular}
}
  \caption{Effects of our augmentation on different state-of-the-art LiDAR-based object detectors on the KITTI validation set. NRConv boosts the performance scores of our method by a large margin in the car detection task. Note that since VirConv~\cite{VirConv} is particularly trained only for detecting cars, the other two classes (pedestrians and cyclists) are omitted. }
\label{objdet_upgrades}
\end{table*}

\begin{table*}[ht]
\resizebox{\textwidth}{!}{%
\begin{tabular}{l||l|l|l|l|l|l|l|l}
Method      & Modality  & \multicolumn{3}{c}{Car 3D AP (R40)}              & \multicolumn{3}{|c|}{Car BEV AP (R40)}             & Time        \\ 
&    & Easy           & Mod.           & Hard           & Easy           & Mod.           & Hard           &             \\ \hline \hline
PV-RCNN \cite{Shi_2020_CVPR}     & LiDAR     & 90.25          & 81.43          & 76.82          & 94.98          & 90.65          & 86.14          & 80          \\
Voxel-RCNN \cite{deng2021voxel}  & LiDAR     & 90.90          & 81.62          & 77.06          & 94.85          & 88.83          & 86.13          & 40          \\
CT3D \cite{Sheng_2021_ICCV}       & LiDAR     & 87.83          & 81.77          & 77.16          & 92.36          & 88.83          & 84.07          & 70          \\
SE-SSD \cite{Zheng_2021_CVPR}     & LiDAR     & 91.49          & 82.54          & 77.15          & 95.68          & \red{91.84}          & 86.72          & \textbf{30} \\
BtcDet \cite{Xu_Zhong_Neumann_2022}     & LiDAR     & 90.64          & 82.86          & 78.09          & 92.81          & 89.34          & 84.55          & 90          \\
CasA \cite{9870747}        & LiDAR     & 91.58          & 83.06          & \textbf{80.08}          & 95.19          & 91.54          & 86.82          & 86          \\
Graph-Po \cite{10.1007/978-3-031-20074-8_38}    & LiDAR     & \red{91.79}          & 83.18          & 77.98          & \red{95.79}          & \textbf{92.12}          & \red{87.11}          & 60          \\
 3ONet \cite{hoang20233onet}   & LiDAR     &  \textbf{92.03}          & \textbf{85.47}          & 78.64          & \textbf{95.87}          & 90.07          & 85.09          & 60          \\\hline \hline 

Aug-VirConv (Ours)   & LiDAR & 90.53 & \red{83.84} & \red{79.10} & 94.52 & 91.00 & \textbf{88.08} & 92
\end{tabular}
}
\caption{Quantitative Results on the KITTI test set. For the sake of fairness, we show LIDAR-only methods. The best results are marked in bold, and the second-best results are  coloured red.}
\label{testset}
\end{table*}

\begin{table}[!b]
    \centering
\resizebox{\columnwidth}{!}{%
  \begin{tabular}{|l|l|l|l|l|l|l|l|l|}
    \hline
     & Modality
    \multirow{2}{*}{} &
      \multicolumn{3}{c|}{Car 3D}  
\\

      \hline
   &  & Easy & Med & Hard      \\
    \hline
    PointPillars \cite{lang2019pointpillars}& LIDAR & 88.28 & 79.32 & 76.72  \\ \hline
    SECOND \cite{yan2018second} & LIDAR &  88.11 & 78.15 & 73.23  \\ \hline 
    PV-RCNN \cite{shi2020pv} & LIDAR & 91.46 & 84.12 & 82.22  \\ \hline
    Voxel-RCNN \cite{deng2021voxel} & LIDAR & 92.38 & 85.29 & 82.86 \\ \hline 
    SE-SSD \cite{Zheng_2021_CVPR} & LIDAR & 93.19 & 86.12 & 83.31   \\ \hline
    BtcDet \cite{Xu_Zhong_Neumann_2022} & LIDAR & 93.15 & 86.28 & 83.86  \\ \hline
    3ONet \cite{hoang20233onet} & LIDAR & \textbf{94.24} & 87.32 & 84.17 \\ \hline
    Aug-VirConv (Ours) & LIDAR &  92.62 & \textbf{87.76} & \textbf{85.34}  \\ \hline

  \end{tabular}
}
  \caption{Results on the KITTI validation set for the LiDAR-only method for the average precision with 40 recall thresholds.}
  \label{validationset}
\end{table}


\subsection{Quantitative Results}

Table~\ref{objdet_upgrades} shows the obtained quantitative results for three different models (PointPillars~\cite{Lang_2019_CVPR}, SECOND~\cite{yan2018second}, and Voxel-RCNN~\cite{deng2021voxel}) trained with and without our augmentation framework on the KITTI validation set. 
For each model, we report the overall precision together with the individual mean average precision (mAP) scores for the car, pedestrian, and cyclists classes across all difficulty strata (easy, medium, and hard). 

This table conveys that all these networks benefit from the dense pseudo-point clouds generated by our framework.
Each of these detectors showcases a performance boost.
For instance, in the case of overall medium difficulty, we obtain $+1.2\%$ and $+1.3\%$ performance upgrade for PointPillars~\cite{Lang_2019_CVPR} and SECOND~\cite{yan2018second}, respectively.
Note that since VirConv~\cite{VirConv} is particularly trained only for detecting cars, the other two classes (pedestrian and cyclists) are omitted in Table~\ref{objdet_upgrades}.
When we particularly focus on the car medium scores, we observe that although there are slight improvements for the PointPillars~\cite{Lang_2019_CVPR} and SECOND~\cite{yan2018second}, the performance upgrade on car detection substantially increases up to  $+2.9\%$ in the case of introducing Noise Resistant Convolution (NRConv) in VirConv~\cite{VirConv} to the Augmented Voxel-RCNN (Ours). 
This clearly shows that models specifically designed to address the noise of pseudo point clouds (e.g., VirConv~\cite{VirConv}) benefit more from our proposed technique than any other model.

We evaluate our method on the KITTI test set and compare it with the state-of-the-art models on the leaderboard.
Table~\ref{testset} reports obtained results for the Car 3D and BEV detection benchmarks. In this experiment, we augment  VirConv~\cite{VirConv} with our pseudo-LiDAR data and obtain comparable results with the other LiDAR-only detectors.
We see that Aug-VirConv (Ours) sets a new state-of-the-art result (\textbf{$88.08\%$}) in the Car BEV Hard difficulty, whereas it achieves the second-best score (\textbf{$83.84\%$}) for the Car 3D Medium difficulty.

We also compare the performances of the LiDAR-only methods on the KITTI validation set. The results in Table \ref{validationset} show that our method is the best in medium and hard difficulty levels. 
The performance drop of our approach in the test set shows that the generalization power of our framework needs to be improved. Our method is limited by the performances of TiTAN-Next~\cite{2023arXiv230207661C} and will benefit from improvements in the multi-modal domain translation research area.

\begin{table}[!b]
\centering
\begin{tabular}{|p{25mm}|p{13mm}|p{15mm}|p{13mm}|}
    \hline
    
    & StVD~\cite{VirConv} & SGP (Ours) & 3D AP    \\
    \hline
    \centering LiDAR points (Voxel-RCNN~\cite{deng2021voxel}) & No & No & 84.9 \\ \hline
    \centering Early Fusion & Yes \newline No & No \newline Yes & 85.6 \newline \textbf{85.7}  \\ \hline
    \centering Late Fusion & Yes \newline No & No \newline Yes & 86.8 \newline \textbf{87.8} \\ \hline
  \end{tabular}
  
\caption{Ablation study on the KITTI validation set for discarding pseudo points on the car-only detection. The goal is to compare StVD \cite{VirConv} with SGP.}
\label{ablation}
\end{table}

\begin{figure}[!b]
\centering
\resizebox{1\columnwidth}{!}{
\includegraphics[]{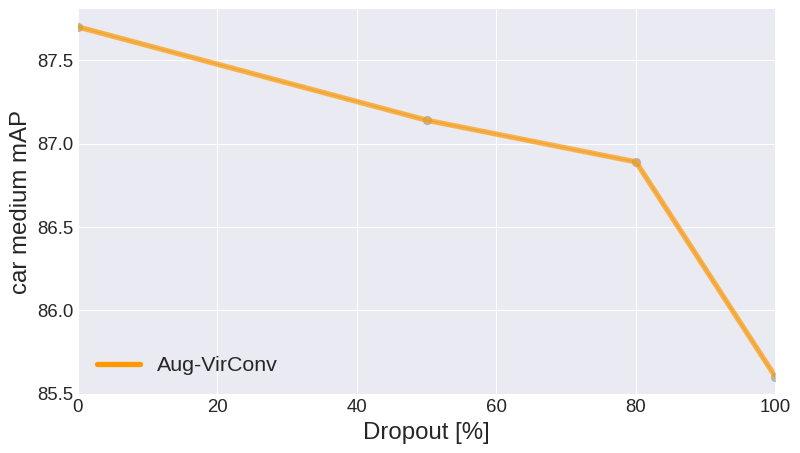}
}
\caption{Study of the input discard rate in our framework. We observe that after applying SGP, dropping more points decreased detection scores.}
\label{inputdisc}
\end{figure}

\subsection{Ablation study}

\paragraph{Influence of the projection method}
VirConv~\cite{VirConv} generates a pseudo point cloud and then employs Stochastic Voxel Discard (StVD) to drop about $80\%$ of the point cloud at every pass. 
Our Semantic Guided Projection (SGP) method already selects only the points with relevant semantic information.
To validate the utility of our pseudo point discard method, we repeat the 3D car detection experiment while switching between StVD and SGP methods.
%
The results are reported in Table~\ref{ablation}. 
Using Voxel-RCNN only on the LiDAR point cloud yields a performance of 84.9 mAP. When we adopt an early fusion approach, that is we concatenate the pseudo point cloud with the real LiDAR, StVD and SGP yield comparable result, with ours being better by 0.1 mAP. When we adopt the late fusion approach of VirConv-T, however, we can see that our method outperforms the stochastic approach of VirConv. Late fusion with StVD brings a 1.9 mAP improvement to the baseline, while our method brings a 2.9 mAP improvement for the medium difficulty of the car detection problem.

\paragraph{Input discard influence}
The pipeline of VirConv~\cite{VirConv} sets a discard of $80\%$ of the voxels at the input of the model. Our method already drops more than $80\%$ of the points while applying our projection scheme  SGP, thus, having an input discard is less relevant in our setting. 
Nevertheless, we investigate whether having additional discard could improve the performances by serving as a data augmentation method. The results are plotted in Figure~\ref{inputdisc}. We can see that the best result is already achieved when there is no discard. This proves that the projection scheme SGP  removes redundant points and there is no need for an additional discard.

\paragraph{Multimodal data}
Our framework follows a LiDAR-only fashion while making use of a multi-modal domain translator (i.e., TITAN-Next~\cite{2023arXiv230207661C}) to create a pseudo point cloud from synthetic depth and semantic segmentation images. 
This pipeline can also work in a multimodal setting by replacing the synthetic images with real images obtained from a real camera.
For this purpose, we use the images from the KITTI dataset and generate the depth and semantic segmentation maps using the off-the-shelf models MIDAS~\cite{Ranftl2022} and SD-Net~\cite{2019arXiv190710659O}, respectively. Next, we employ SGP to generate the pseudo point cloud. 
The results are reported in Table~\ref{multimodal}.
VirConv~\cite{VirConv} adds the real RGB information to the pseudo point cloud while generating the point cloud. To have a fair comparison, we train the VirConv model without RGB values and compare the results with this version. We observe that for the multimodal setting, SGP and StVD have comparable performances, with slightly better results for StVD~\cite{VirConv}. One reason for that may be that our method cumulates the error from both the depth prediction and the semantic segmentation, while VirConv has depth prediction error, only. We believe that advances in semantic segmentation will bridge the  performance gap between these two methods. 
Moreover, compared to VirConv~\cite{VirConv} our method has the benefit of being non-stochastic. StVD randomly drops $80\%$ of the pseudo points at the input stage, which for a single frame may result in dropping the most important points and a loss in performances. However, our method is deterministic since we select the points based on semantic segmentation data. Thus, for the same frame, we always get the same input point cloud. This means that the performance is more stable and the reasons for failure are easier to trace in real world applications.



\begin{table}[]
\centering
\resizebox{\columnwidth}{!}{
\begin{tabular}{|l|l|l|l|l|l|l|l|l|}
    \hline
     & Mod
    \multirow{2}{*}{} &
      \multicolumn{3}{c|}{Car 3D} & 
      \multicolumn{3}{c|}{Car BEV} 
\\
      \hline
   &  & Easy & Med & Hard & Easy & Med & Hard     \\
    \hline
    VirConv-T$^*$~\cite{VirConv} & LC & 92.8 & \textbf{88.3} & \textbf{87.8} & \textbf{96.4} & \textbf{93.4} & \textbf{91.3} \\ \hline
    Augmented VirConv-T$^*$ (ours) & LC & \textbf{95.0} & 88.0 & 86.7 & 95.5 & 93.2 & 91.1 \\ \hline

  \end{tabular}%
    
}%

  \caption{Results obtained when training in a multimodal setting. VirConv-T$^*$~\cite{VirConv} represents the model trained without the RGB values. For a fair comparison, real camera segment and depth maps are added to our augmentation framework.}
  \label{multimodal}
\end{table}

\section{Discussion and Conclusion}

In this work, we introduce a novel semantics-aware LiDAR-only pseudo-point cloud generation for 3D Object Detection.
We leverage the scene semantic and depth information coming from the multi-modal domain translation module TITAN-Next without requiring any additional sensor modality, such as cameras.  
The final output of our framework is a denser LiDAR point cloud with semantic segments and 3D bounding boxes for some specific objects. 

To the best of our knowledge, this proposed framework is the first of its kind that returns augmented and semantically segmented LiDAR scans without camera sensors.

%
Reported experimental results showed that our framework is agnostic to object detector architectures, however, works with higher performance in the case of having detectors specifically designed for pseudo point cloud data, such as VirConv~\cite{VirConv}.
\\

The main limitation of the proposed framework is the computation time (see Table~\ref{testset}) needed to create the synthetic depth and semantic segmentation maps. Since each module forward passes is quite fast, an efficient pipeline that reduces dead time can mitigate this issue. For example, TITAN-Next could process the next frame while the detection module is processing the current point cloud. \\

\bibliographystyle{IEEEtran} 
\bibliography{refernces}

\begin{thebibliography}{10}
\providecommand{\url}[1]{#1}
\csname url@samestyle\endcsname
\providecommand{\newblock}{\relax}
\providecommand{\bibinfo}[2]{#2}
\providecommand{\BIBentrySTDinterwordspacing}{\spaceskip=0pt\relax}
\providecommand{\BIBentryALTinterwordstretchfactor}{4}
\providecommand{\BIBentryALTinterwordspacing}{\spaceskip=\fontdimen2\font plus
\BIBentryALTinterwordstretchfactor\fontdimen3\font minus
  \fontdimen4\font\relax}
\providecommand{\BIBforeignlanguage}[2]{{%
\expandafter\ifx\csname l@#1\endcsname\relax
\typeout{** WARNING: IEEEtran.bst: No hyphenation pattern has been}%
\typeout{** loaded for the language `#1'. Using the pattern for}%
\typeout{** the default language instead.}%
\else
\language=\csname l@#1\endcsname
\fi
#2}}
\providecommand{\BIBdecl}{\relax}
\BIBdecl

\bibitem{Lang_2019_CVPR}
A.~H. Lang, S.~Vora, H.~Caesar, L.~Zhou, J.~Yang, and O.~Beijbom,
  ``Pointpillars: Fast encoders for object detection from point clouds,'' in
  \emph{Proceedings of the IEEE/CVF Conference on Computer Vision and Pattern
  Recognition (CVPR)}, June 2019.

\bibitem{mao2021pyramid}
J.~Mao, M.~Niu, H.~Bai, X.~Liang, H.~Xu, and C.~Xu, ``Pyramid r-cnn: Towards
  better performance and adaptability for 3d object detection,'' \emph{ICCV},
  2021.

\bibitem{Sheng_2021_ICCV}
H.~Sheng, S.~Cai, Y.~Liu, B.~Deng, J.~Huang, X.-S. Hua, and M.-J. Zhao,
  ``Improving 3d object detection with channel-wise transformer,'' in
  \emph{Proceedings of the IEEE/CVF International Conference on Computer Vision
  (ICCV)}, October 2021, pp. 2743--2752.

\bibitem{Shi_2019_CVPR}
S.~Shi, X.~Wang, and H.~Li, ``Pointrcnn: 3d object proposal generation and
  detection from point cloud,'' in \emph{Proceedings of the IEEE/CVF Conference
  on Computer Vision and Pattern Recognition (CVPR)}, June 2019.

\bibitem{9018080}
S.~Shi, Z.~Wang, J.~Shi, X.~Wang, and H.~Li, ``From points to parts: 3d object
  detection from point cloud with part-aware and part-aggregation network,''
  \emph{IEEE Transactions on Pattern Analysis and Machine Intelligence},
  vol.~43, no.~8, pp. 2647--2664, 2021.

\bibitem{Yang_2020_CVPR}
Z.~Yang, Y.~Sun, S.~Liu, and J.~Jia, ``3dssd: Point-based 3d single stage
  object detector,'' in \emph{Proceedings of the IEEE/CVF Conference on
  Computer Vision and Pattern Recognition (CVPR)}, June 2020.

\bibitem{Yang_2019_ICCV}
Z.~Yang, Y.~Sun, S.~Liu, X.~Shen, and J.~Jia, ``Std: Sparse-to-dense 3d object
  detector for point cloud,'' in \emph{Proceedings of the IEEE/CVF
  International Conference on Computer Vision (ICCV)}, October 2019.

\bibitem{8886046}
D.~Zhou, J.~Fang, X.~Song, C.~Guan, J.~Yin, Y.~Dai, and R.~Yang, ``Iou loss for
  2d/3d object detection,'' in \emph{2019 International Conference on 3D Vision
  (3DV)}, 2019, pp. 85--94.

\bibitem{wang2019pseudo}
Y.~Wang, W.-L. Chao, D.~Garg, B.~Hariharan, M.~Campbell, and K.~Weinberger,
  ``Pseudo-lidar from visual depth estimation: Bridging the gap in 3d object
  detection for autonomous driving,'' in \emph{CVPR}, 2019.

\bibitem{you2020pseudo}
Y.~You, Y.~Wang, W.-L. Chao, D.~Garg, G.~Pleiss, B.~Hariharan, M.~Campbell, and
  K.~Q. Weinberger, ``Pseudo-lidar++: Accurate depth for 3d object detection in
  autonomous driving,'' in \emph{ICLR}, 2020.

\bibitem{10.1007/978-3-030-58601-0_19}
X.~Ma, S.~Liu, Z.~Xia, H.~Zhang, X.~Zeng, and W.~Ouyang, ``Rethinking
  pseudo-lidar representation,'' in \emph{Computer Vision -- ECCV 2020},
  A.~Vedaldi, H.~Bischof, T.~Brox, and J.-M. Frahm, Eds.\hskip 1em plus 0.5em
  minus 0.4em\relax Cham: Springer International Publishing, 2020, pp.
  311--327.

\bibitem{VirConv}
H.~Wu, C.~Wen, S.~Shi, and C.~Wang, ``Virtual sparse convolution for multimodal
  3d object detection,'' in \emph{CVPR}, 2023.

\bibitem{10.1007/978-3-030-64559-5_16}
T.~Cortinhal, G.~Tzelepis, and E.~Erdal~Aksoy, ``Salsanext: Fast,
  uncertainty-aware semantic segmentation of lidar point clouds,'' in
  \emph{Advances in Visual Computing}, G.~Bebis, Z.~Yin, E.~Kim, J.~Bender,
  K.~Subr, B.~C. Kwon, J.~Zhao, D.~Kalkofen, and G.~Baciu, Eds.\hskip 1em plus
  0.5em minus 0.4em\relax Cham: Springer International Publishing, 2020, pp.
  207--222.

\bibitem{2023arXiv230207661C}
T.~{Cortinhal} and E.~{Erdal Aksoy}, ``{Depth- and Semantics-aware Multi-modal
  Domain Translation: Generating 3D Panoramic Color Images from LiDAR Point
  Clouds},'' \emph{arXiv e-prints}, p. arXiv:2302.07661, Feb. 2023.

\bibitem{lang2019pointpillars}
A.~H. Lang, S.~Vora, H.~Caesar, L.~Zhou, J.~Yang, and O.~Beijbom,
  ``Pointpillars: Fast encoders for object detection from point clouds,'' in
  \emph{Proceedings of the IEEE/CVF conference on computer vision and pattern
  recognition}, 2019, pp. 12\,697--12\,705.

\bibitem{yan2018second}
Y.~Yan, Y.~Mao, and B.~Li, ``Second: Sparsely embedded convolutional
  detection,'' \emph{Sensors}, vol.~18, no.~10, p. 3337, 2018.

\bibitem{deng2021voxel}
J.~Deng, S.~Shi, P.~Li, W.~Zhou, Y.~Zhang, and H.~Li, ``Voxel r-cnn: Towards
  high performance voxel-based 3d object detection,'' in \emph{Proceedings of
  the AAAI Conference on Artificial Intelligence}, vol.~35, no.~2, 2021, pp.
  1201--1209.

\bibitem{vora2020pointpainting}
S.~Vora, A.~H. Lang, B.~Helou, and O.~Beijbom, ``Pointpainting: Sequential
  fusion for 3d object detection,'' in \emph{Proceedings of the IEEE/CVF
  conference on computer vision and pattern recognition}, 2020, pp. 4604--4612.

\bibitem{xu2021fusionpainting}
S.~Xu, D.~Zhou, J.~Fang, J.~Yin, Z.~Bin, and L.~Zhang, ``Fusionpainting:
  Multimodal fusion with adaptive attention for 3d object detection,'' in
  \emph{2021 IEEE International Intelligent Transportation Systems Conference
  (ITSC)}.\hskip 1em plus 0.5em minus 0.4em\relax IEEE, 2021, pp. 3047--3054.

\bibitem{wang2021pointaugmenting}
C.~Wang, C.~Ma, M.~Zhu, and X.~Yang, ``Pointaugmenting: Cross-modal
  augmentation for 3d object detection,'' in \emph{Proceedings of the IEEE/CVF
  Conference on Computer Vision and Pattern Recognition}, 2021, pp.
  11\,794--11\,803.

\bibitem{huang2020epnet}
T.~Huang, Z.~Liu, X.~Chen, and X.~Bai, ``Epnet: Enhancing point features with
  image semantics for 3d object detection,'' in \emph{Computer Vision--ECCV
  2020: 16th European Conference, Glasgow, UK, August 23--28, 2020,
  Proceedings, Part XV 16}.\hskip 1em plus 0.5em minus 0.4em\relax Springer,
  2020, pp. 35--52.

\bibitem{yin2021multimodal}
T.~Yin, X.~Zhou, and P.~Kr{\"a}henb{\"u}hl, ``Multimodal virtual point 3d
  detection,'' \emph{Advances in Neural Information Processing Systems},
  vol.~34, pp. 16\,494--16\,507, 2021.

\bibitem{wu2022sparse}
X.~Wu, L.~Peng, H.~Yang, L.~Xie, C.~Huang, C.~Deng, H.~Liu, and D.~Cai,
  ``Sparse fuse dense: Towards high quality 3d detection with depth
  completion,'' in \emph{Proceedings of the IEEE/CVF Conference on Computer
  Vision and Pattern Recognition}, 2022, pp. 5418--5427.

\bibitem{Cortinhal_2021_ICCV}
T.~Cortinhal, F.~Kurnaz, and E.~E. Aksoy, ``Semantics-aware multi-modal domain
  translation: From lidar point clouds to panoramic color images,'' in
  \emph{Proceedings of the IEEE/CVF International Conference on Computer Vision
  (ICCV) Workshops}, October 2021, pp. 3032--3048.

\bibitem{yin2019fusionmapping}
P.~Yin, J.~Qian, Y.~Cao, D.~Held, and H.~Choset, ``Fusionmapping: Learning
  depth prediction with monocular images and 2d laser scans,'' \emph{arXiv
  preprint arXiv:1912.00096}, 2019.

\bibitem{kittidataset}
A.~Geiger, P.~Lenz, and R.~Urtasun, ``Are we ready for autonomous driving? the
  kitti vision benchmark suite,'' in \emph{2012 IEEE Conference on Computer
  Vision and Pattern Recognition}, 2012, pp. 3354--3361.

\bibitem{Shi_2020_CVPR}
S.~Shi, C.~Guo, L.~Jiang, Z.~Wang, J.~Shi, X.~Wang, and H.~Li, ``Pv-rcnn:
  Point-voxel feature set abstraction for 3d object detection,'' in
  \emph{Proceedings of the IEEE/CVF Conference on Computer Vision and Pattern
  Recognition (CVPR)}, June 2020.

\bibitem{Zheng_2021_CVPR}
W.~Zheng, W.~Tang, L.~Jiang, and C.-W. Fu, ``Se-ssd: Self-ensembling
  single-stage object detector from point cloud,'' in \emph{Proceedings of the
  IEEE/CVF Conference on Computer Vision and Pattern Recognition (CVPR)}, June
  2021, pp. 14\,494--14\,503.

\bibitem{Xu_Zhong_Neumann_2022}
\BIBentryALTinterwordspacing
Q.~Xu, Y.~Zhong, and U.~Neumann, ``Behind the curtain: Learning occluded shapes
  for 3d object detection,'' \emph{Proceedings of the AAAI Conference on
  Artificial Intelligence}, vol.~36, no.~3, pp. 2893--2901, Jun. 2022.
  [Online]. Available:
  \url{https://ojs.aaai.org/index.php/AAAI/article/view/20194}
\BIBentrySTDinterwordspacing

\bibitem{9870747}
H.~Wu, J.~Deng, C.~Wen, X.~Li, C.~Wang, and J.~Li, ``Casa: A cascade attention
  network for 3-d object detection from lidar point clouds,'' \emph{IEEE
  Transactions on Geoscience and Remote Sensing}, vol.~60, pp. 1--11, 2022.

\bibitem{10.1007/978-3-031-20074-8_38}
H.~Yang, Z.~Liu, X.~Wu, W.~Wang, W.~Qian, X.~He, and D.~Cai, ``Graph r-cnn:
  Towards accurate 3d object detection with semantic-decorated local graph,''
  in \emph{Computer Vision -- ECCV 2022}, S.~Avidan, G.~Brostow, M.~Ciss{\'e},
  G.~M. Farinella, and T.~Hassner, Eds.\hskip 1em plus 0.5em minus 0.4em\relax
  Cham: Springer Nature Switzerland, 2022, pp. 662--679.

\bibitem{hoang20233onet}
H.~A. Hoang and M.~Yoo, ``3onet: 3d detector for occluded object under
  obstructed conditions,'' \emph{IEEE Sensors Journal}, 2023.

\bibitem{shi2020pv}
S.~Shi, C.~Guo, L.~Jiang, Z.~Wang, J.~Shi, X.~Wang, and H.~Li, ``Pv-rcnn:
  Point-voxel feature set abstraction for 3d object detection,'' in
  \emph{Proceedings of the IEEE/CVF conference on computer vision and pattern
  recognition}, 2020, pp. 10\,529--10\,538.

\bibitem{Ranftl2022}
R.~Ranftl, K.~Lasinger, D.~Hafner, K.~Schindler, and V.~Koltun, ``Towards
  robust monocular depth estimation: Mixing datasets for zero-shot
  cross-dataset transfer,'' \emph{IEEE Transactions on Pattern Analysis and
  Machine Intelligence}, vol.~44, no.~3, 2022.

\bibitem{2019arXiv190710659O}
M.~{Ochs}, A.~{Kretz}, and R.~{Mester}, ``{SDNet: Semantically Guided Depth
  Estimation Network},'' \emph{arXiv e-prints}, p. arXiv:1907.10659, Jul. 2019.

\end{thebibliography}

\end{document}